# Multichannel consecutive data cross-extraction with 1DCNN-attention for diagnosis of power transformer

1st Wei Zheng
State Key Lab of Electrical
Insulation and Power Equipment
Xi'an Jiaotong University
Xi'an, China
e-mail:
zhengwei990927@stu.xjtu.edu.cn

2nd Guogang Zhang
State Key Lab of Electrical
Insulation and Power Equipment
Xi'an Jiaotong University
Xi'an, China
e-mail:
ggzhang@mail.xjtu.edu.cn

3rd Chenchen Zhao
State Key Lab of Electrical
Insulation and Power Equipment
Xi'an Jiaotong University
Xi'an, China
e-mail:
cc.zhao@stu.xjtu.edu.cn

4th Qianqian Zhu
State Key Lab of Electrical
Insulation and Power Equipment
Xi'an Jiaotong University
Xi'an, China
e-mail:
qianqianzhu@stu.xjtu.edu.cn

*Abstract*—Power transformer plays a critical role in grid infrastructure, and its diagnosis is paramount for maintaining stable operation. However, the current methods for transformer diagnosis focus on discrete dissolved gas analysis, neglecting deep feature extraction of multichannel consecutive data. The unutilized sequential data contains the significant temporal information reflecting the transformer condition. In light of this, the structure of multichannel consecutive data cross-extraction (MCDC) is proposed in this article in order to comprehensively exploit the intrinsic characteristic and evaluate the states of transformer. Moreover, for the better accommodation in scenario of transformer diagnosis, one dimensional convolution neural network attention (1DCNN-attention) mechanism is introduced and offers a more efficient solution given the simplified spatial complexity. Finally, the effectiveness of MCDC and the superior generalization ability, compared with other algorithms, are validated in experiments conducted on a dataset collected from real operation cases of power transformer. Additionally, the better stability of 1DCNN-attention has also been certified.

## I. Introduction

Due to the essential function, the facilities of power transformer are crucial for power system. Considering the high frequency of use and expensive maintenance costs, the precise state identification for power transformer is significant to keep the stability and reliability of grid [1][2]. Oil-immersed transformer, as the most widely implemented transformer category, are insulated by mineral oil. However, the long-term consecutive operation may result in the thermal and electrical defects, which could lead to server failure [3]. Currently, the dissolved gas analysis (DGA) has become a main index for oil-immersed transformer condition evaluation due to the close relationship and mature monitoring technology [4].

In recent decades, with the fast development of perception technology (PT) and data mining (DM), there are various methods based on DGA are putted forward and implemented for oil-immersed transformer diagnosis. Currently, they can be sorted into rule-based and ML-based methods. The rule-based methods attempt to establish a set of standard according to the mathematical statistic, representatively, such as Rogers ratio method, IEC three ratio method, Dornenburg method and Duval Triangle method [5][6][7][8]. As far as ML-based methods, with the popularization of pattern recognition in various scenarios, more and more neural networks are proposed with different backbones and analysis angles. For example, in 2018, Muthi A analyzed the transformer states and detected incipient faults using artificial neural networks (ANN) [9]. In 2017, Dai used deep belief network (DBN) to analyzing the relationship between the dissolved gases in dielectric oil and states of transformer [10]. In 2022, Elsisi proposed one-dimension convolutional neural network (1D-CNN) due to the robustness against uncertainties [11]. The aforementioned methods all verify that the perception data of DGA obtains the characteristics reflecting to the transformer state.

Generally speaking, the transformer diagnosis is an evolutionary application of pattern recognition. With the presence of more complex task of pattern recognition involved more modalities and dimensional traits, various deep learning (DL) methods have been proposed, such as Segment Anything Model (SAM) of Meta AI Lab [12], Attention Mechanism (AM) of Google [13] and Mixer of HUAWEI Ltd [14] to realize natural image segment (NIS), natural language understanding (NLD) and multivariate time series forecasting (MTSF). Additionally, automatic facial expression recognition (AFER) [15] and speech recognition (SR) [16] are also achieved. Meanwhile, more and more DL mechanism are introduced into facility diagnosis with the fair performances. Yang proposed a polynomial kernel induced distance metric to modify the deep transfer learning for machines diagnosis [17]. Prieto proposed a neural network to promise bearing degradation detection [18]. Zhao proposed a deep residual shrinkage network to achieve a cracking feature learning ability from highly noised vibration signals [19]. Based on generative adversarial network (GAN), Zhou healed with unbalanced data for fault diagnosis [20]. In 2021, Zheng proposed an improved fusion single shot multi-box to detect the insulation state of substation insulator according to the infrared image [21]. In summary, even these methods have different backbones, for the same purpose of connecting the diverse information or monitoring signals with different patterns, there is a consistent core mind of them, which is obtaining the significant features.

As far as hidden features extraction, the AM have been certified as an effective way. The AM can be grouped into two categories: cross-attention and self-attention. Even their sources of Query are different, all of them have achieved significant success in various field. Such as native language comprehension

[13], Ocean Current Prediction [22], Computer Vison (CV) [23], and Video Salient Target Detection [24] and so on. In general, the AM obtains the internal priorities and correlations of inputs by using learning matrixes, which means that there is a large number of parameters need to be learned.

One of the main advantages of CNN network is parameter sharing to realize small parameter models (SPM). Currently, the implementations of CNN have achieved an extraordinary success in fault diagnosis, such as the deep convolutional neural network (DCNN) for compound fault diagnosis of rotating machinery [25]. Additionally, as a special form of CNN, 1DCNN not only has fewer parameters but also has been verified being more natural to deal with 1-D signal for diagnosis [26]. In the scenario of DGA, even there are various signals (multichannel) of gases are involved, each gas concentration is collected and conveyed as 1-D data stream. Moreover, when the timing is taken as the object, the combination of each gas is also 1-D data.

Despite current advances, most recent researches of transformer diagnosis are all based on the discrete DGA. But the discrete data could contain a larger proportion of randomness during the field-collection. In addition, there are same researches have proved the identity of DGA to evaluate the dynamic transition process of power transformer condition [27][28]. The timing sequences of DGA can not only reduce the interference of sampling noise but also involve the temporal adjacent information. Therefore, in order to contain more perceived information of transformer condition, the consecutive dissolved gases data (CDGD) is collected and finally the 2-D map of CDGD is formed as the index for transformer diagnosis in this article.

When CDGD is defined as the research object, the more critical core is how to extract the multidimensional characteristics hidden in CDGD. Although there are many existing feature extraction algorithms, their performances are often difficult to guaranteed in distinct scenarios. The self-attention mechanism, due to the outstanding generalization and rationality, has attracted much attention. However, the attention map, as the core component of AM, can only describes the attention distribution in one dimension of the original input. But the CDGD involved temporal and channel dimensions. Each dimensions includes the indispensable feature. It can be understood that, under different transformer conditions, not only some gases are more significant but also some inter-temporal changes are more valuable for reference. Therefore, the structure of Multichannel consecutive data cross-extraction (MCDC) is proposed to successively extract the different dimension-wise features, which describes the emphasis and relevance among different moments and diverse channels.

Additionally, due to the conventional self-attention depends on learning matrix parameters, there is a dilemma of that the algorithmic space complexity geometrically increases as the input scale increase. When the task does not need to mine overly complex spaces, the global optimal convergence would be quite difficult. Therefore, this article proposed an 1DCNN-attention mechanism, which fuses the advantage of CNN weight sharing with attention idea. Moreover, the better efficiency and naturalization of 1DCNN in processing 1D data are also exploited when evaluating the attention map in one dimension of CDGD.

In a nutshell, for the purpose to sustain power transformer long-term reliable running, this article tends to achieve a better performance of transformer diagnosis with the additional temporal traits contained in CDGD and proposed the MCDC structure and 1DCNN-attention mechanism for more comprehensive feature extraction and better fitness. Finally, the confirmatory experiments are carried out. The core content of this article can be depicted as following sections:

(1) The structure of MCDC and mechanism of 1DCNN-attention are proposed, the details of them are interpreted and the significances are described.
(2) The different structural hyperparameters of MCDC are studied to demonstrate the diverse effect on the diagnostic performance.
(3) The MCDC is compared with some discretization methods to verify the idea of more plentiful indexes contained in temporal sequence.
(4) The MCDC is compared with other continuity algorithms in effectiveness and generalization to prove the superiority of MCDC.
(5) The experimental results and the underlaying meanings are discussed in detail. Furtherly, the potential problems of MCDC application in other scenarios are also expected and discussed.

## II. PROPOSED METHOD

### A. Multichannel consecutive data cross-extraction (MCDC)

The core mind of MCDC is extracting the significant information on time sequence and channel successively. The overall architecture of the MCDC is depicted in Fig. 1

The multichannel sequential data, including hydrogen ($H_2$), methane ($CH_4$), ethane ($C_2H_6$), ethylene ($C_2H_4$) and acetylene ($C_2H_2$), is inputted into the framework. There are four following modules in MCDC: embedding, temporal interaction, channel interaction and projection. In case of feature disappearing during forward process, the residual structure is added between modules. The overall procedure of MCDC as follow:

$$X_h^E = \text{embed}(X_{input}) \quad (1)$$

$$X_h^T = \text{temporal}(X_h^E) \quad (2)$$

$$X_h^C = \text{channel}(X_h^T + X_h^E) \quad (3)$$

$$Y_{output} = \text{project}(X_h^C + X_h^T) \quad (4)$$

Where the $X_{input} \in \mathbb{R}^{5 \times Temporal}$ is the original input map of consecutive gases concentration, $Temporal$ depends on the temporal length of the CDGD, which is set as 8 in the early phase of this article. $Y_{output} \in \mathbb{R}^{1 \times V}$ depicts the probability distribution, $V$ depends on the variety of transformer condition.

In the Embedding module, the sinusoidal positional encoding [14] is added into the original feature map to emphasize the timeliness of each channel. This procedure can be demonstrated as follow:

$$\text{embed}(X_{input}) = X_{input} + \text{PE}(X_{input}) \quad (5)$$

$$\begin{cases} PE(X_{input})_{(2i,t)} = \sin(t/(10000^{2i/d_{channel}})) \\ PE(X_{input})_{(2i+1,t)} = \cos(t/(10000^{2i/d_{channel}})) \end{cases} \quad (6)$$

Where the $PE(X_{input}) \in \mathbb{R}^{5 \times Temporal}$ is the sinusoidal position map, which dimensions equals to $X_{input}$. The $d_{channel}$ is the number of channels of 5 and $t = (0,1,...,Temporal-1)$.

In the module of Temporal Interaction, the 1DCNN-attention mechanism is implemented for extracting the information at separate timing and mapping the temporal-wise attention map to emphasize the relevance between different moments. Additionally, the structure of Multi-head Attention is laid out to capture richer representation of information feature, according to more attention maps from distinct subspaces. This procedure can be depicted as follow:

$$\begin{cases} T_i = \text{attention}(X_h^E, X_h^E, X_h^E) \\ T = \text{Concat}(T_1, T_2, \cdots, T_H) \\ \text{temporal}(X_h^E) = M^T \otimes T \end{cases} \quad (7)$$

Where $i = (1,2,...H)$. $H$ is the number of head and the $T_i \in \mathbb{R}^{n \times Temporal}$ is the output of $i$th temporal-attention layer, $n$ depends on the number of channels, size of kernel, number of padding and stride in temporal interaction. The $M^T \in \mathbb{R}^{5 \times hn}$ is set for formation of residual.

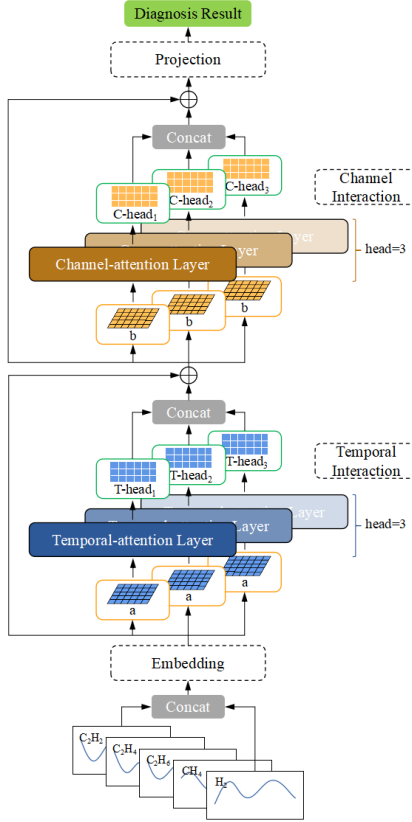

Fig. 1. The overall architecture of MCDC

In the module of Channel Interaction, the 1DCNN-attention mechanism is employed again. But the difference is convolution object. The purpose of channel interaction is extracting the information on individual channels and mapping the channel-wise attention map to underline the correlation between different channels. In this paper, each channel is the concentration of each gas. Muti-head Attention is also used for its superiority. This procedure as follow:

$$\begin{cases} C_i = \text{attention}(X_h^T + X_h^E, X_h^T + X_h^E, X_h^T + X_h^E) \\ C = \text{Concat}(C_1, C_2, \cdots, C_H) \\ \text{channel}(X_h^T + X_h^E) = C \otimes M^C \end{cases} \quad (8)$$

Where $H$ is the number of head and the $C_i \in \mathbb{R}^{5 \times m}$ is the output of $i$th Channel-attention layer, $m$ depends on the length of data, size of kernel, number of padding and stride in channel interaction. The $M^C \in \mathbb{R}^{hm \times Temporal}$ is set for formation of residual.

Finally, the projection module outputs the diagnosis result through feedforward neural network (FFN) with activation function of sigmoid and softmax layer. As follow:

$$\text{project}(X_h^C + X_h^T) = \text{Softmax}(FFN(X_h^C + X_h^T)) \quad (9)$$

### B. 1DCNN-attention

In continuous data, there are local priorities and internal correlations. The self-attention mechanism can extract and focus on that by calculating the vectors of Query, Key and Value. Conventional attention mechanism is to learn respective matrices to obtain these vectors. Even though this method has promising performance for large-scale natural language understanding, its complex algorithmic space increases the probability of getting stuck with local optimum, especially when the task does not need to mine overly complex spaces. Therefore, the structure of 1DCNN-attention is proposed in this section to apply with the task of transformer diagnosis, which detail is depicted in Fig. 2. The Fig. 2 demonstrates the detailed processes and data structure in Temporal Interaction.

The input of gas features map is convolved by three different convolution kernels, distinguished by different colours of yellow (on the top layer), green (on the middle layer) and red (on the bottom layer), to acqurie the Query, Key and Value vectors. The specific kernel size and way of padding and stride is adjustable. The process of getting Query, Key and Value can be defined as:

$$\begin{cases} Q = \text{Convol}_Q(input) \\ K = \text{Convol}_K(input) \\ V = \text{Convol}_V(input) \end{cases} \quad (10)$$

Furtherly, generating the similarity or correlation between Query and Key as attention map $\in \mathbb{R}^{Temporal \times Temporal}$, which can be understood as the inner wight coefficient of the Value. The formula for attention map can be described as:

$$Amap = \text{Softmax}(\frac{K^T \otimes Q}{\sqrt{T}}) \quad (11)$$

Finally, the weighted summation of Value gives the output of attention-paid features map as follow:

$$output = \text{attention}(input) = V \otimes Amap \quad (12)$$

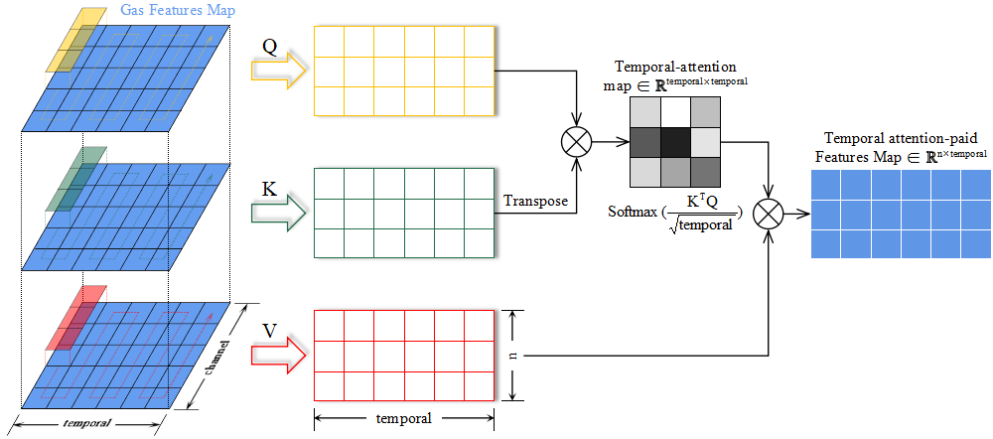

Fig. 2. The 1DCNN-attention of Temporal Interaction

## C. Cross-extraction

The idea of 1DCNN-attention mechanism is extracting multi-dimensional features in a certain dimension of the information map and using similarity to emphasize the coherence between tokens of that dimension. However, the existence of local priorities and internal correlations could be in diverse dimensions. Such as the information map of CDGD in task of transformer diagnosis, there are two dimensions of temporal and channel. Among the moments, there are key timings and temporal development which indexes different condition of transformer. Among the channels, the emphasis on certain gases and gas ratios is also significant in state diagnosis.

In order to extracting and using as much the effective features of information map as possible, the conception of Cross-extraction is proposed and applied into MCDC as the modules of Temporal Interaction and Channel Interaction. In the implementation of Cross-extraction for MCDC, the major distinction in modules is the convolution route, shown as Fig. 3. In Fig. 3, the time span of CDGD is 6 days. The lower areas covered in blue (left) and dark yellow (right) are the input feature maps for Temporal Interaction and Channel Interaction respectively and the grey areas present the padding tokens. The upper red area means a 1D-convolution kernel with size of 3 and the dotted arrow line shows the path of convolution.

In Temporal Interaction, the kernel extracts the feature crossing channels in a certain timing and gets Query, Key and Value of Temporal Interaction $\in \mathbb{R}^{n \times Temporal}$ for the Temporal-wise attention map $\in \mathbb{R}^{Temporal \times Temporal}$. In Channel Interaction, the feature extracted is crossing timing in a certain channel. The Query, Key and Value of Channel Interaction $\in \mathbb{R}^{5 \times m}$ and the Channel-wise attention map $\in \mathbb{R}^{5 \times 5}$.

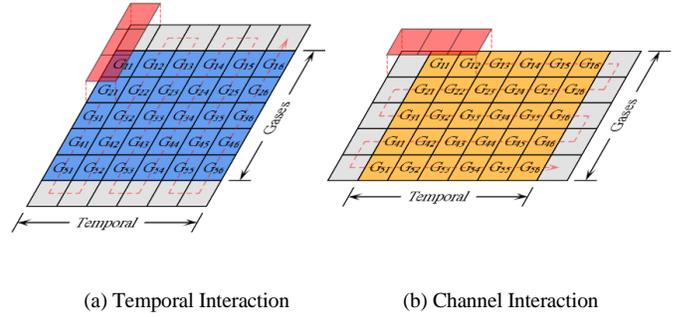

(a) Temporal Interaction      (b) Channel Interaction

Fig. 3. Cross-extraction for MCDC

## III. EXPERIMENT AND ANALYSIS

Python 3.6 with pytorch 1.9.0 is applied to run the experiment code. The experimental hardware conditions are i5-11400F CPU and NVIDIA GeForce GTX 3060 GPU. All algorithms operate on the same dataset and platform.

### A. Dataset and experiment setup

In order to identify the method proposed in this article, the CDGD of 65 power transformers in operation or under inspection are collected from Book-Typical Cases [29][30][31] and Research Literatures [28][32]. The voltage levels of transformers include 35kV, 110kV, 220kV and 500kV to match the realistic data distribution. The conditions of transformer involve normal condition (NC), low overheating (LT), medium overheating (MT), high overheating (HT), partial discharge (PD), low energy discharge (LD) and high energy discharge (HD), which are all indexed according to the CDGD. Fig. 4 illustrates the distribution of transformers under varying conditions as well as their temporal spans of the perceptive CDGD. The sampling frequency of the CDGD is once per day and Multiple interpolation method is employed to address data gaps resulting from sampling deficiency. The Fig. 5 demonstrates the curves of CDGD. Due to the space limitation of this article, only one of the 500kV transformers in MT is showcased. The nonlinearity of each gas curves is obvious. Nevertheless, in some local areas, there is a certain degree of similarity between different gas trends, while some show opposite property.

For the purpose of enhancing data, the process of overlapping sampling is taken to expand the numbers of the dataset and enrich the variety of temporal features in the data, shown as Fig. 6. The TABLE I. presents a detail overview of the dataset's structural composition. For instance, within the dataset, there are 8 transformers pertaining to NC, exhibiting temporal span ranges from 24 to 205 days. After overlapping sampling with *Temporal* of 8, a cumulative total of 421 CDGD samples of NC were obtained.

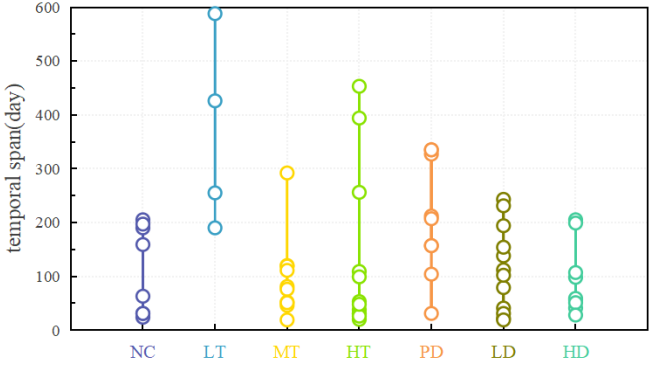

Fig. 4. CDGD distribution

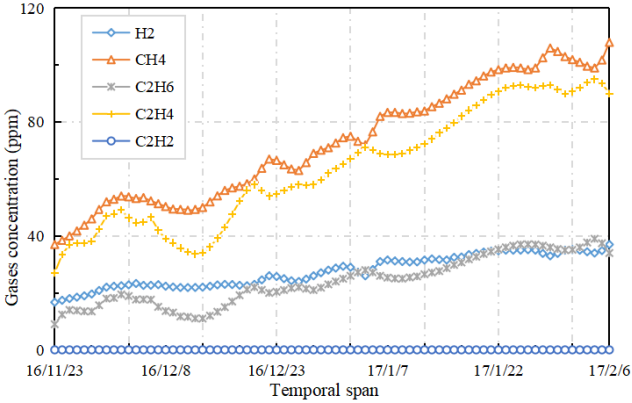

Fig. 5. CDGD curves of MT

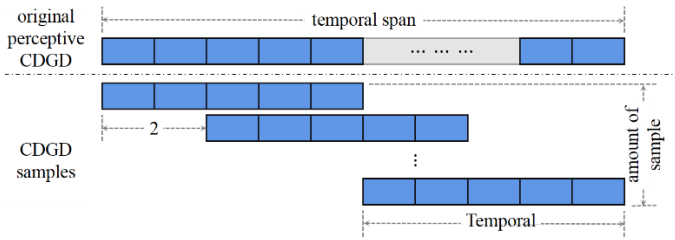

Fig. 6. Overlapping sampling

TABLE I. THE STRUCTURE OF DATASET

| Condition (Encoding) | Amount of Transformer | Amount of CDGD sample |
|---|---|---|
| NC (0) | 8 | 421 |
| LT (1) | 4 | 714 |
| MT (2) | 10 | 446 |
| HT (3) | 12 | 733 |
| PD (4) | 10 | 1064 |
| LD (5) | 12 | 638 |
| HD (6) | 9 | 389 |

The dataset is divided stochastically into 80% of training set and 20% of testing set. In order to avoid differences in distribution of data trait, the crossfold validation approach with 4 folds is chosen for training MCDC. The entire amount of training epoch is 1000. Ealy interruption mechanism is involved for avoiding overfitting. Loss function of cross-entropy is selected for calculating the reverse gradient, and optimizer of Adam is chosen for learning parameters of MCDC. The original learning rate is 0.01, which decays twice, to 0.001 at 500th epoch and to 0.0002 at 750th epoch for digging the extremity. The batch size is set as 200. The main assessment indexes including Accuracy, Precision, Recall and F1-value. Their formulas are described as follow:

$$Ac = (TN + TP)/(TN + TP + FN + FP) \quad (13)$$

$$Pr = TP/(TP + FP) \quad (14)$$

$$Re = TP/(TP + FN) \quad (15)$$

$$F1 = 2 \times (Pr \times Re)/(Pr + Re) \quad (16)$$

Where, the TN means the true negative amount, the TP means the true positive amount, the FN means the false negative amount and FP means the false positive amount.

### B. Structure hyperparameters analysis

Different hyperparameters of network structure are compared and analyzed in this subsection in order to study how is the performance of MCDC affected by them. The main structural hyperparameters include the kernel sizes of Temporal Interaction and Channel Interaction as well as the number of heads for Multi-head Attention. For the sake of exploitation of realistic information and reduction of introduced information, the convolution stride is set as 1 and the way of padding is to ensure that output has the same shape as input.

Fig. 7 demonstrates the map of accuracy with different conjugate kernel size. The mean projected bands of single kernel size are also shown at the top and side of Fig. 7 to reveal the significance of single kernel size in different Interaction. It can be known that the kernel size of Temporal Interaction is 5 giving the best performance when it is only considered. This is quite apprehensible from the point of view of the convolution receptive field. The convolution of Temporal Interaction extracts the features crossing channels and there are five channels in total. Therefore, the kernel of size 5 can extract more comprehensive features by better coverage. The best kernel size of Channel Interaction is 6 instead of 8 to equal with *T*. The information redundancy between some of the different timing nodes can be the cause. Considering the accuracy map doesn't take other structural hyperparameters into consideration, the kernel size of Temporal Interaction of 5 and kernel size of Channel Interaction of 6 are chosen for the rest experiments.

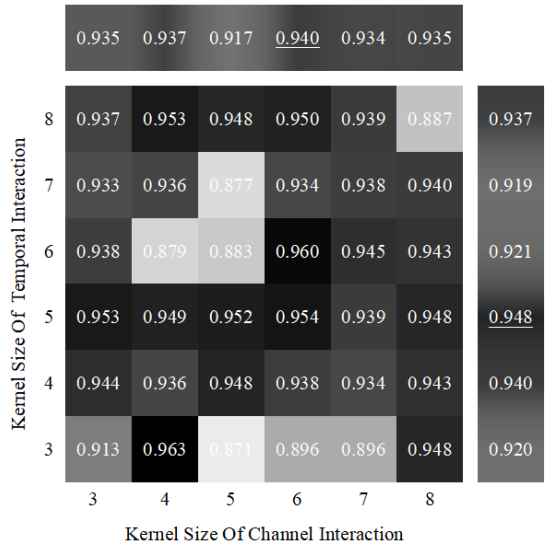

Fig. 7. Accuracy map of different kernel size

Fig. 8 depicts the accuracy with different number of heads. At the same time, the 1DCNN-attention (convolution-based) mechanism is compared with conventional matrix-based attention mechanism. Since the main distinction is reflected in the algorithmic stability, repetition test is carried out by 10 times and the indexes of maximum mean error and mean accuracy are introduced. It is obvious that, whatever the number of heads is, the mean accuracy of 1DCNN-attention exceeds that of matrix-based by a margin of 0.57% at least. This margin is augmented to 1.1% as the number of head increases. And the performance of 1DCNN-attention is more stable when the number of heads increases. Furthermore, the variation tendency of accuracy with number of heads is same, no matter it is matrix-based or convolution-based. This can prove the effectiveness of multi-head mechanism in emphasizing the internal manifestation form more perspectives. Considering the 4-heads achieves the best accuracy, the heads number of 4 is chosen for the rest experiments.

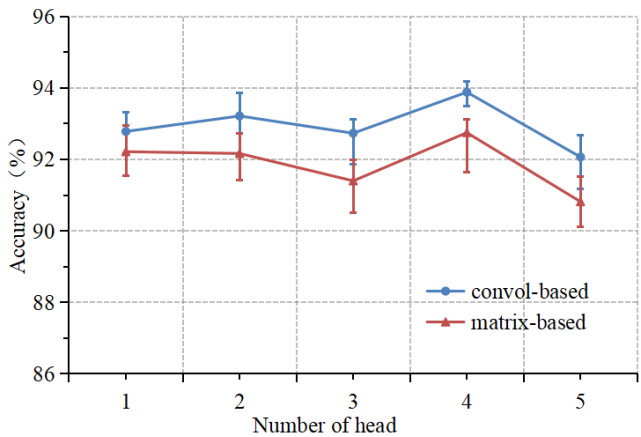

Fig. 8. Accuracy of different number of heads

In the scenario of transformer diagnosis, the temporal length of CDGD data is also a significant index. Fig. 9 demonstrates the accuracy of MCDC with different temporal length. Obviously, the performance of MCDC gets increasement as the length increases. Whereas, lengthening data leads to shrinking volume of dataset and augment in computing resources required. In the rest experiments, the temporal length is set as 12 due to the best performance.

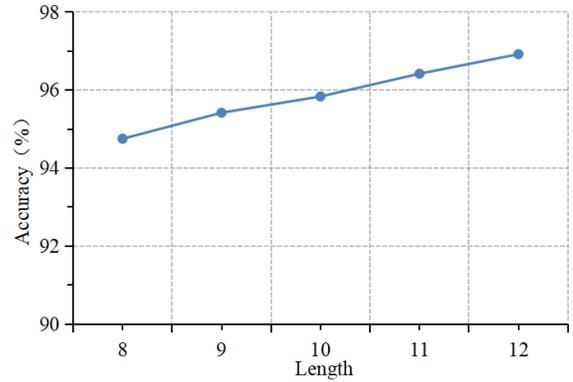

Fig. 9. Accuracy of different data length

### C. The training process of MCDC

In this subsection, cross validation with 4 folders is carried out for training MCDC. The loss curve is shown as Fig. 10 and validation accuracy curve is shown as Fig. 11. For the purpose of verifying the data distribution variance, the loss-crossfold and accuracy-crossfold are also depicted.

With the rapid decline of loss value, the accuracy of validation rises quickly in the early 100 epochs. In the following 400 epochs, the performance keeps getting better gradually and circuitously. Especially, the network acquires a steady improvement again when the learning rate decays at $500^{th}$ epoch. It proves that learning rate decay can exploit the potential of model furtherly. Macroscopically, the curves of crossfold have the same tend. Therefore, the variance of data is not apparent. To sum up, MDCD can learns effective knowledge to suit the task of transformer diagnosis with the help from Adam optimizer.

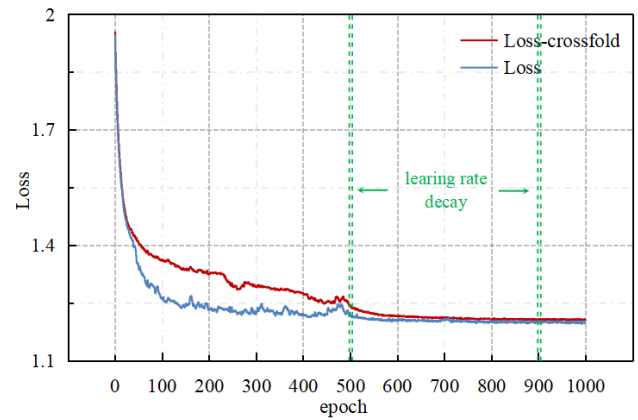

Fig. 10. Loss curve of MCDC during training process

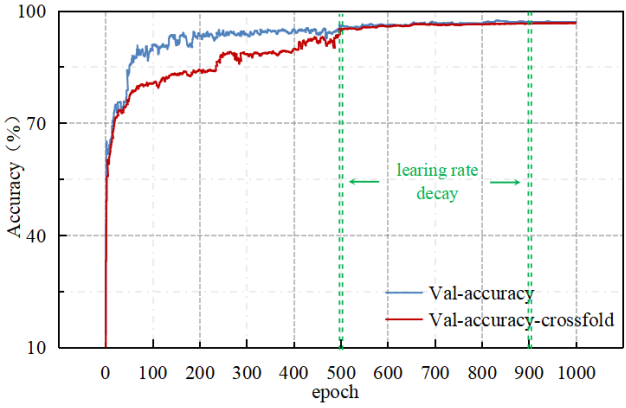

Fig. 11. Validation accuracy curve of MCDC during training process

In order to evaluate the advantage and effectivity of method proposed in this article, the confusion matrix is established in this part as a recognized approach, shown as Fig. 12. The rows of confusion matrix mean the real conditions and columns mean the classify result from MCDC. It can be known that the MCDC possesses full-score performance at condition 'HT'. Even there is relative weakness in conditions 'NC' and 'LD', the results of other conditions are all higher than 0.97. In general, the structure of MCDC acquires an ideal performance on the task of transformer diagnosis.

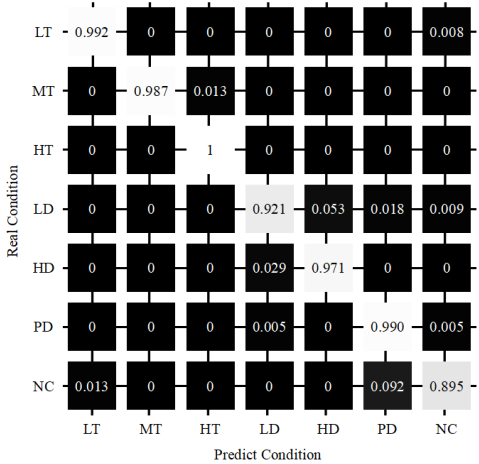

Fig. 12. Confusion matrix of MCDC

### D. Comparison with discretization methods

In this subsection, some transformer diagnostic algorithms dealing with discrete data are tested for the comparison with method proposed in this article. The conventional artificial neural network (ANN) [9], energy-weighted dissolved combustible gases (EWDCG) [33], conventional convolution neural network (CNN) [11] are involved. ANN consists of 3 layers of feedforward network (FFN) with activation function of sigmoid. EWDCG relies the enthalpies of formation of each gas to weight them and introduces rates of change per month of each gas as additive indexes. CNN extracts the feature of input and outputs the possibilities of each class through a fully connected layer.

TABLE II. exhibits the diagnostic performances of different methods and bolds the best result. It can be known from TABLE II. that MCDC produce the best result in any indicator. EWDGA performs better than ANN and CNN. The merit of EWDGA can be attributed to containing the information of gases change in time series. The slight enhancement of CNN from ANN proves the effectivity of feature extraction by CNN.

TABLE II. PERFORMANCES OF DIFFERENT DISCRETIZATION METHODS

| Method | Ac | Macro-Pr | Macro-Re | Macro-F1 |
|---|---|---|---|---|
| ANN | 0.913 | 0.908 | 0.883 | 0.887 |
| EWDGA | 0.940 | 0.946 | 0.921 | 0.925 |
| CNN | 0.919 | 0.923 | 0.903 | 0.908 |
| MCDC | *0.971 | *0.970 | *0.965 | *0.967 |

a. Asterisk indicates the best result

In order to demonstrate the distinct performances of different algorithms in more detail, the indicators of receiver operating characteristic (ROC) and area under the curve (AUC) [11] are utilized due to the outstanding evaluation ability for binary classification, especially in the scenarios of multiple classification. In addition, the macro-average ROC and micro-average ROC are introduced, as well as their AUCs, to present a global assessment. The ROC curves and AUCs for each class of different methods are shown as Fig. 13. For the ANN, there is relatively obvious weakness in classes of LD and NC, although both are higher than 0.82. EWDGA and CNN both improve the performance oriented to NC, which AUC values all increased above 0.95. Whereas, the flaw of HD is even lightly worse. MCDC achieves the best result while meliorates the drawbacks of HD and LD simultaneously. In terms of binary classification issue, the MCDC shows the perfect result of LT and HT and the AUC of others are all higher than 0.95, with the macro-average AUC of 0.96 and micro-average AUC of 0.96. what can be known is that the feature extraction in CNN and time sequence information in EWDGA are both significant. They also support the best performance produced by MCDC.

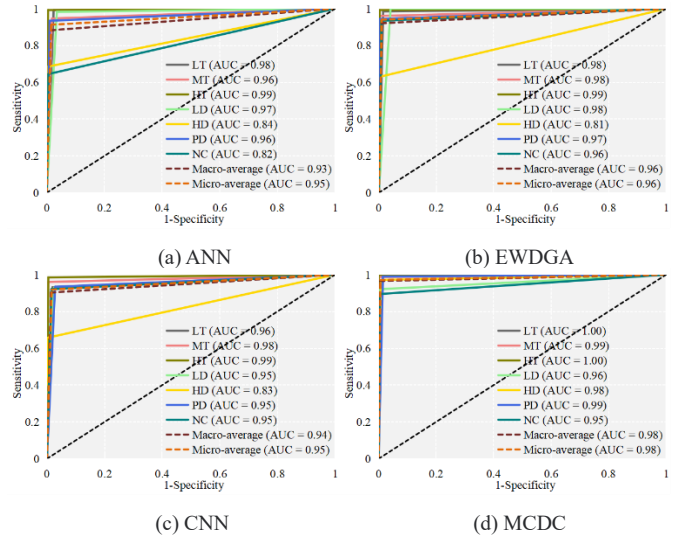

(a) ANN  (b) EWDGA
(c) CNN  (d) MCDC

Fig. 13. Results of ROC curve and AUC

*E. Comparison with continuity algorithms*

In this subsection, the continuous data processing algorithms are implemented for comparison in the scenario of transformer diagnosis, which contains hidden Markov model (HMM) [28], gate recurrent unit (GRU) [34] and deep belief network (DBN) [35]. The HMM establishes the relationship between observed properties, hidden state and observed state and the transition probability matrix of hidden state to describe the sequence. The GRU is a type of recurrent neural network (RNN), which utilizes the reset gate and update gate to learn the transition process of series. DBN is a deep learning algorithm using multi-layers restricted Boltzmann machine (RBM) to extract the traits of input.

TABLE III. exhibits the performances of different diagnosis algorithms. These algorithms can be ranked as MCDC > HMM > GRU > DBN, according to the performance indicator of accuracy. The best result is achieved by MCDC while the worst is obtained by DBN.

TABLE III.    PERFORMANCES OF DIFFERENT CONTINUITY ALGORITHMS

| Algorithm | Ac | Macro-Pr | Macro-Re | Macro-F1 |
|---|---|---|---|---|
| HMM | 0.954 | 0.947 | 0.956 | 0.945 |
| GRU | 0.941 | 0.947 | 0.924 | 0.932 |
| DBN | 0.879 | 0.881 | 0.837 | 0.845 |
| MCDC | *0.971 | *0.970 | *0.965 | *0.967 |

b. Asterisk indicates the best result

For the sake of verifying the generalization of the proposed method. MCDC, HMM, GRU and DBN compare again with different structure of dataset. The sources of dataset do not change. But the dividing object for the training set and the testing set is replaced with the facility. The 65 power transformers are stochastically divided into 50 as training set and 15 as testing set, both covering all varieties of condition, which details are shown in TABLE IV. The results of various algorithms are shown in TABLE V. These algorithms can be ranked as MCDC > GRU > DBN > HMM, according to the indicators. All the algorithms show apparent decline in performance. This is caused by local overfitting of information distribution. But the MCDC still keep the first-class with accuracy of 0.910. The property of HMM turns into the worst from previous second best. This is attributed to the linear reflecting mode of HMM between hidden state and observed properties, which validity would fluctuate greatly when the magnitude of original input changes. The large interval of transformer voltage level takes a huge challenge for algorithmic generalization. It can be known that the generalization ability of deep learning methods of GRU and DBN are much better. However, the MCDC achieves the best.

TABLE IV.    THE DETAILS OF DATASET

| Condition | Amount of Transformer (train) | Amount of CDGD sample (train) | Amount of Transformer (test) | Amount of CDGD sample (test) |
|---|---|---|---|---|
| NC | 6 | 321 | 2 | 84 |
| LT | 3 | 584 | 1 | 122 |
| MT | 8 | 372 | 2 | 54 |
| HT | 9 | 680 | 3 | 29 |
| PD | 8 | 720 | 2 | 324 |
| LD | 9 | 464 | 3 | 150 |
| HD | 7 | 315 | 2 | 56 |

TABLE V.    PERFORMANCES OF GENERALIZATION OF DIFFERENT CONTINUITY ALGORITHMS

| Algorithm | Ac | Macro-Pr | Macro-Re | Macro-F1 |
|---|---|---|---|---|
| HMM | 0.668 | 0.573 | 0.517 | 0.479 |
| GRU | 0.819 | 0.703 | 0.712 | 0.672 |
| DBN | 0.796 | 0.616 | 0.689 | 0.650 |
| MCDC | *0.910 | *0.887 | *0.842 | *0.847 |

c. Asterisk indicates the best result

To prove the significant difference between the algorithms, the Wilcoxon rank-sum test [36] is carried out for a quantificational indicator of p-value, which is used to interpret the difference between two distributions. TABLE VI. exhibits the results of Wilcoxon rank-sum test of different algorithms on the benchmark of MCDC. All the p-values are smaller than 0.05. The hypothesis of differences between diagnostic results can be accepted, which supports the superiority of MCDC than other algorithms.

TABLE VI.    P-VALUES OF DIFFERENT ALGORITHMS

| Algorithm | P-value |
|---|---|
| HMM | 3.15E-02 |
| GRU | 2.99E-05 |
| DBN | 3.02E-03 |

IV. DISCUSSION

The core thought of proposed MCDC is extracting potential features in various dimensions of CDGD step by step, in order to better distinguish the different states in the field of power transformer diagnosis. TSNE [37] is a visualization tool which can achieve the clustering dimension reduction to observe the feature distribution of multi-categories data. Fig. 14 exhibits the visualized results of data at different layers of MCDC. It can be known that, when data is extracted through the Temporal Interaction, the data of same state have a cluster tendency while the boundaries between different state are not distinct. When the data of CDGD pass through the whole processes of MCDC, the boundaries are much clearer.

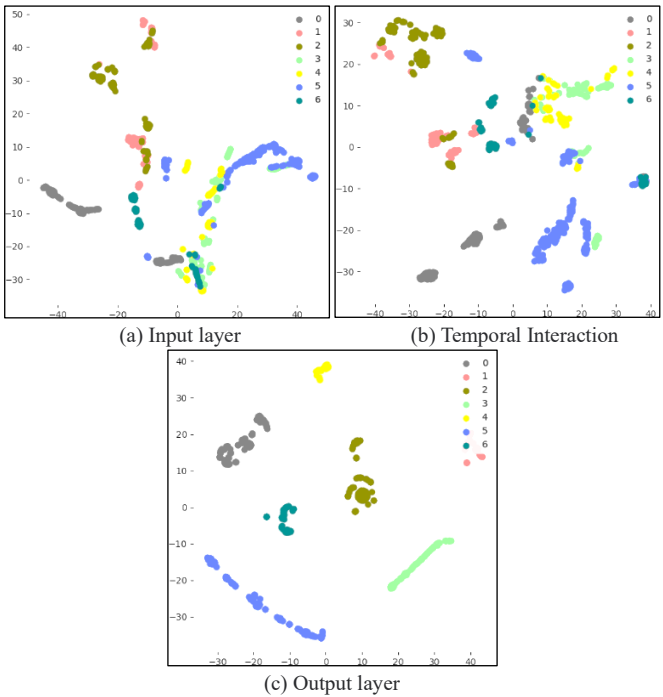

(a) Input layer
(b) Temporal Interaction
(c) Output layer

Fig. 14. The TSNEs of different layers

Additionally, the 1DCNN-attention mechanism, proposed to adjust to the field of diagnosis of transformer, proves to be more stable than conventional matrix-based attention mechanism. Compared with discretization methods, the MCDC outperforms ANN, EWDGA and CNN by enhancing the accuracy with 5.86%, 3.13% and 5.25%. It means the sequence features are significant for strengthening the transformer diagnosis method. Compared with continuity algorithms, the MCDC outperforms HMM, GRU and DBN by enhancing the accuracy with 1.75%, 3% and 9.25%. Whereas, when the dataset is facility-wisely divided, the enhancements increase to 24.25%, 9.13% and 11.38%. The superiority of MCDC and property of deep learning of MCDC can be certified, especially according to the generalization ability, which is quite significant in the scene implements.

Although the MCDC gives a competitive performance. The 1DCNN-attention mechanism maybe limited by the modality of perceptional data and the way of data combination. Therefore, the extraction operation method needs to be adjusted, when different data modalities, such as infrared image and primitive signal with high frequency, are taken into consideration and the performances need to be verified furtherly.

## V. CONCLUSION

In this article, the structure of MCDC is demonstrated for the purpose of comprehensively exploring the potential features of CDGD information map in the various dimensions, thus enhancing diagnostic capability. The 1DCNN-attention mechanism is proposed and integrated into the framework of MCDC, in order to acclimatize to the scenario of power transformer diagnosis. The CDGD dataset is collected from 65 transformers operating at diverse voltage levels and operation conditions. According to the results of verification experiments and comparison experiments, the timing sequential information contained in CDGD can be effectively extracted by MCDC and utilized to improve the performance of transformer diagnosis. Furthermore, the better stability of the 1DCNN-attention is verified than conventional self-attention mechanism. The superior abilities of extraction and generalization of MCDC are also verified. In summary, the MCDC with 1DCNN-attention proposed in this article has the capability to extract the characteristics hiding in the dimensions of channel and temporality and achieves good performance in power transformer diagnosis.

## Declaration of Competing Interest

The authors declare that they have no known competing financial interests or personal relationships that could have appeared to influence the work reported in this paper.

## Acknowledgment

This work was supported by the Science and Technology Project of the State Grid Corporation of China under Grant 5500-202199527A-0-5-ZN.

## Data availability

Data will be made available on request.

## Declaration of generative AI and AI-assisted technologies in the writing process

During the preparation of this work the authors used Chat GPT in order to improve language and readability. After using this tool, the authors reviewed and edited the content as needed and take full responsibility for the content of the publication.